\documentclass[10pt,twocolumn,letterpaper]{article}

\usepackage{cvpr}
\usepackage{times}
\usepackage{epsfig}
\usepackage{graphicx}
\usepackage{amsmath}
\usepackage{amssymb}
\usepackage{multirow}
\usepackage{booktabs}
\usepackage{subfig}
\usepackage{float}
\usepackage{caption}
\usepackage{rotating}
 
\usepackage[pagebackref=true,breaklinks=true,letterpaper=true,colorlinks,bookmarks=false]{hyperref}

\cvprfinalcopy 

\def\cvprPaperID{2047} 

\ifcvprfinal\fi
\begin{document}

\title{Compressive Image Recovery Using Recurrent Generative Model}

\author{Akshat Dave, Anil Kumar Vadathya, Kaushik Mitra\\
{\tt\small \{ee12b073,ee15s055,kmitra\}@ee.iitm.ac.in}\\
Indian Institute of Technology, Madras\\
}

\makeatletter
\let\@oldmaketitle\@maketitle
\renewcommand{\@maketitle}{\@oldmaketitle
\begin{center}
    \vspace{-0.0cm}
    \includegraphics[trim=0 18 0 8,clip, width=1\textwidth]{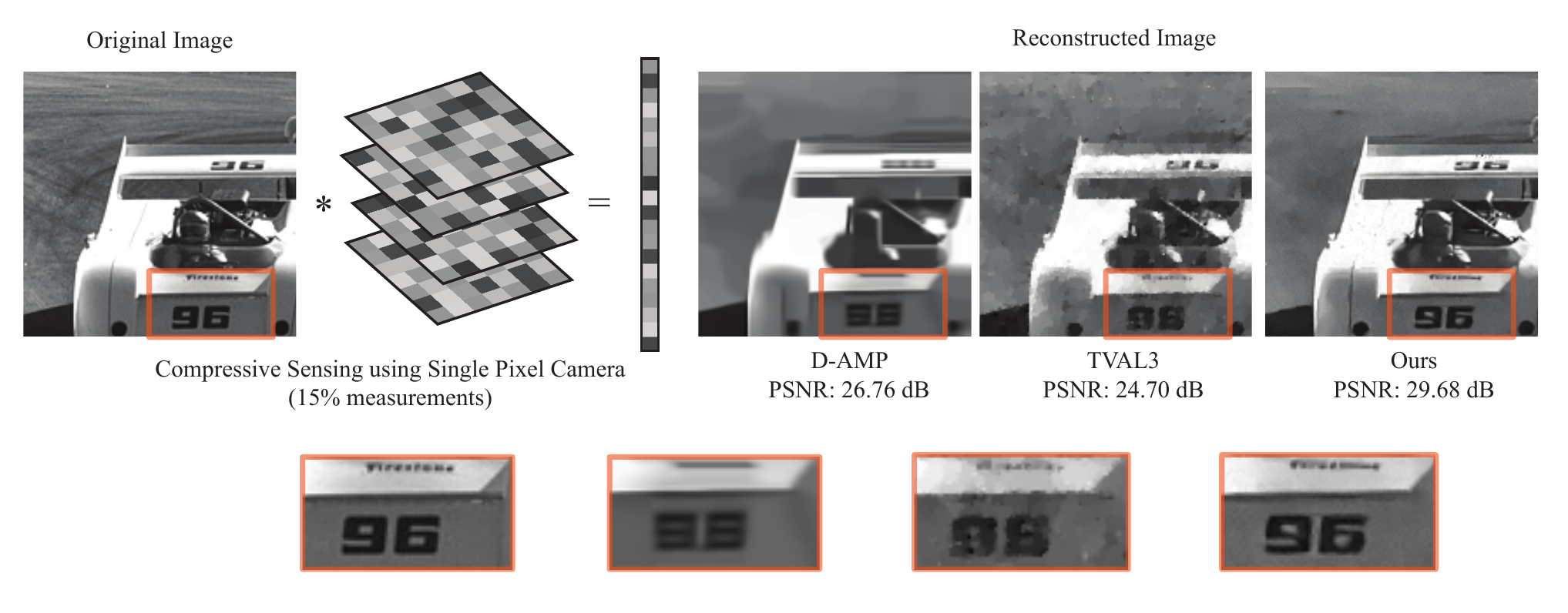}
    \captionof{figure}{We propose to use a deep generative model, RIDE \cite{theis2015generative} , as an image prior for compressive signal recovery. Since RIDE models long-range dependency in images using spatial LSTM, we are able to recover the image better than other competing methods.}
    \label{fig:fig1}
    \vspace{-0.0cm}
\end{center}
\bigskip}
\makeatother

\maketitle

\begin{abstract}

Reconstruction of signals from compressively sensed measurements is an ill-posed problem. In this paper, we leverage the recurrent generative model, RIDE, as an image prior for compressive image reconstruction. Recurrent networks can model long-range dependencies in images and hence are suitable to handle global multiplexing in reconstruction from compressive imaging.  We perform MAP inference with RIDE using back-propagation to the inputs and projected gradient method. We propose an entropy thresholding based approach for preserving texture in images well. Our approach shows superior reconstructions compared to recent global reconstruction approaches like D-AMP and TVAL3 on both simulated and real data. 

\end{abstract}

\section{Introduction}
\label{sec:intro}
Imaging in the non-visible region of the spectrum has a plethora of applications owing to its unique properties  \cite{hansen2008overview}. For example, improved penetration of infrared waves through fog and smog enables imaging through scattering media. However, prohibitive sensing costs in the non-visible range have limited its widespread use\footnote{Megapixel sensors in short-wave infrared,  typically constructed using InGaAs, cost more than USD 100k.}. Many works have proposed Compressive Sensing (CS) \cite{baraniuk2007compressive, donoho2006compressed} as a viable solution for high-resolution imaging beyond the visible range of spectrum \cite{duarte2008single, sankaranarayanan2012cs, chen2015fpa}. Compressive sensing theory states that signals exhibiting sparsity in some transform domain can be reconstructed from much lower measurements than sampling at Nyquist rate \cite{donoho2006compressed}. Lesser the number of measurements lesser is the cost of sensing. The single-pixel camera (SPC) is a classical example of CS framework \cite{duarte2008single}. In SPC, a single photo diode is used to capture compressive measurements and then reconstruct back the whole scene. 


A challenge faced by CS reconstruction algorithms is to recover a high dimensional signal from a small number of measurements. This ill-posed nature of the reconstruction makes data priors essential. Often, signals exhibit sparse structure in some transform domain. For example, natural images in the domain of wavelets, DCT coefficients or gradients. Initially, reconstruction methods exploited this prior knowledge about the signal structure thereby restricting the solution set to desired signal \cite{li2013efficient,sankaranarayanan2012cs,chen2015fpa,wang2015lisens}. However, using these simple sparsity based priors at very low measurement rates results in low-quality reconstructions (see TVAL3 reconstruction in fig. \ref{fig:fig1}). This is due to their inability to capture the complexity of natural image statistics. On the other hand, data-driven approaches have been proposed recently to handle the complexity \cite{aghagolzadeh2012compressive, kulkarni2016reconnet,mousavi2015deep}. They led to successful results in terms of reconstruction. But all of these approaches handle only local multiplexing i.e measurements are taken from image patches and recovery is also done patch wise. This is not appealing for classical SPC framework as such since measurements are acquired through global multiplexing. 

To address these problems, in this work we propose to use a data-driven global image prior, RIDE, proposed by Theis et al. \cite{theis2015generative} for CS image recovery. RIDE uses recurrent networks with Long Short Term Memory (LSTM) units and is shown to model the long-term dependencies in images very well. Also, being recurrent it is not limited to patch size, hence can handle the global multiplexing in SPC. Our contributions are as follows:
\begin{itemize}
  \item We propose to use RIDE as an image prior to model long-term dependencies for reconstructing compressively sensed images. 
  \item We use backpropagation to inputs while doing gradient ascent for MAP inference.
  \item We hypothesize that the model's uncertainty in prediction can be related to the entropy of component posterior probabilities. By thresholding the entropy, we enhance texture preserving the ability of the model. 
\end{itemize}

\section{Prior Work}
\textbf{Role of Signal Priors}: Image data priors have played a significant role for signal reconstruction from ill posed problems which are very common in image processing and computational photography. Initially such image priors were constructed through empirical observation of data statistics, for example TV norm minimization, sparse gradient prior \cite{levin2007deconvolution} and sparsity of coefficients in wavelet domain \cite{portilla2003image}. On the other hand, many methods were proposed to learn the priors directly from data such as dictionary learning \cite{aharon2006img}, mixture models like GMMs \cite{zoran2011learning} and their variants GSMs \cite{portilla2003image}, conditional models like Mixture of Conditional GSMs (MCGSM) \cite{theis2012mixtures}, undirected models like Field of Experts (FoEs) \cite{roth2005fields}.
In dictionary learning an overcomplete set of basis is learnt by representing natural image patches as sparse linear combination of these basis. It has been successfully applied for many image processing tasks \cite{mairal2008learning, aharon2006img}.  On the contrary, rest of the approaches explicitly model the data distribution by maximizing likelihood. GMMs are quite popular image patch priors and have been used for restoration tasks like image denoising and deblurring \cite{zoran2011learning} where it gives competitive results compared to state-of-the-art methods like BM3D \cite{dabov2009bm3d} and KSVD \cite{aharon2006img}. FoEs \cite{roth2005fields} is another popular model which is a Product of Experts (PoEs) with the desirable property of translational invariance making it a whole image prior. It has been used for image inpainting and denoising. 

\textbf{Deep Nets for Image Processing}: Many recent approaches have been proposed to use feed forward deep networks for image reconstruction problems. Burget et al. \cite{burger2012image} used Multilayer perceptrons (MLP) for image patch denoising performing on par with BM3D.  Mao et al. \cite{mao2016image} used very deep convolutional encoder-decoder with skip connections for image denoising even handling different levels of Gaussian noise. It has surpassed BM3D's performance. Xu et al. \cite{xu2014deep} have used convnets for image deconvolution.  Kulkarni et al. \cite{kulkarni2016reconnet} trained a convnet, termed as ReconNet, to recover image from compressed measurements of image patches with measurements being as low as $1\%$. Although these feed forward discriminative models are very fast at run time, their application is limited to the task they are trained for.  Burget et al. \cite{burger2012image} reported difficulty in generalizing a MLP network trained at a particular noise level for different levels of Gaussian noise. Mao et al. \cite{mao2016image} handle this but at the cost of a huge network. ReconNet proposed for CS signal recovery requires the network to be trained again and again for each different sensing matrix and at each different measurement rate. 

\textbf{Deep Generative Models}: Owing to the inherent problems posed by discriminative models, recently much effort has gone into building generative models such as, Generative Adversarial Nets (GAN) \cite{goodfellow2014generative}, Variational Auto Encoders (VAE) \cite{kingma2013auto}, Pixel Recurrent Neural Networks (PixelRNN) \cite{van2016pixel} and Recurrent Image Density Estimator (RIDE) \cite{theis2015generative}. GANs learn the ability to generate a plausible sample from the distribution of natural images. VAE provides a probabilistic framework for both encoding data to latent representation and decoding from it. Auto regressive models like RIDE model the current pixel distribution conditioned on the causal context where Spatial Long Short Term Memory (SLSTM) \cite{graves2012neural} units are used to obtain the contextual summary. PixelRNN is also an auto regressive model like RIDE but with much more complex architecture achieving the state-of-the-art performance in terms of loglikelihood scores. Apart from being expressive, RIDE and PixelRNN come with added advantages. Their directed nature facilitates the computation of exact likelihood. Also, these priors being auto regressive aren't limited to patch size, as is the case with discriminative and even non deep generative models. This is very useful particularly in cases like single pixel camera where the reconstruction has to take account of global multiplexing and patch based methods can't be used directly. 

Among these deep generative models we find RIDE particularly suitable as low level image prior for our tasks involving Bayesian inference. GANs don't model the data distribution and VAE doesn't provide the exact likelihood. PixelRNN although models the distribution, it discretizes the distribution of a pixel to 256 intensity values resulting in optimization difficulties. In this work we extend RIDE as an image prior for reconstruction problems in compressive sensing and image inpainting. 

\textbf{Inpainting}: Image inpainting has been previously attempted with image priors. FoEs were applied to remove scratches or unwanted effects like text from an image. Theis et al. \cite{theis2012mixtures} used conditional model MCGSM for image inpainting. Dictionary learning \cite{mairal2008sparse} has also been proposed for image inpainting although not ideal since it is patch based. A multiscale adaptive version of dictionary learning  \cite{mairal2008learning} is shown to perform well.

 \textbf{Single Pixel Camera}: SPC \cite{duarte2008single} is a compressive sensing framework  \cite{candes2006robust}, where the goal is to reconstruct the image back from a very less number of random linear measurements. 
 Typically this is an ill-posed problem and hence we need to use signal priors. Initially algorithms were proposed to minimize the $l_1$ norm assuming sparsity in the domain of wavelet coefficients, DCT coefficients or gradients \cite{li2013efficient}. Later class of algorithms known as approximate message passing (AMP) algorithms \cite{donoho2009message} \cite{metzler2014denoising} use off-the-shelf denoiser to iteratively refine their solution. ReconNet is another recent method using CNNs. But it can only handle local multiplexing since it is a patch based approach. Here we propose to do compressive image reconstruction with recurrent generative model RIDE as the image prior. Since it is not patch limited, we can handle global multiplexing. 

\section{Background}
Let $\mathbf{x}$ be a gray scale-image and $x_{ij}$ be the pixel intensity at location $ij$ then $\mathbf{x}_{<ij}$ describes the causal context around that pixel containing all $x_{mn}$ such that $m\leq i$ and $j<n$. Now the joint distribution over the image can be factorized as follows: 
\begin{equation}
  p(\mathbf{x}) = \prod_{ij} p(x_{ij}|\mathbf{x}_{<ij}, \boldsymbol{\theta}_{ij})
 \label{factorization}
\end{equation}
where $\boldsymbol{\theta}_{ij}$ are distribution parameters at that location. By making the Markov assumption we can limit the extent of $\mathbf{x}_{<ij}$ to a smaller neighbourhood. Another valid assumption is stationarity of the data which results in sharing the same parameters $\boldsymbol{\theta}$ across all locations $ij$, thus achieving translational invariance. 

Now each factor in the above equation can be modeled by a mixture of GSMs with shared parameters $\boldsymbol{\theta}$ which makes it Mixture of Conditional Gaussian Scale Mixtures (MCGSM) as proposed by \cite{theis2012mixtures}, 
\begin{equation}
  p(x_{ij}|\mathbf{x}_{<ij},\boldsymbol{\theta}) = \sum_{c,s} p(c,s|\mathbf{x}_{<ij},\boldsymbol{\theta})p(x_{ij}|\mathbf{x}_{<ij},c,s,\boldsymbol{\theta}), 
  \label{mcgsm}
\end{equation}
Where the sum is over components and scales, 
\begin{eqnarray}
  p(c,s|\mathbf{x}_{<ij}) &\propto& \exp( \eta_{cs} - 0.5* e^{\alpha_{cs}}\mathbf{x}_{<ij}^{T}\mathbf{K}_c\mathbf{x}_{<ij}),  \nonumber \\ 
  p(x_{ij}|\mathbf{x}_{<ij},c,s) &=&  \mathcal{N}(x_{ij};\mathbf{a}_c^T\mathbf{x}_{<ij}, e^{-\alpha_{cs}})  \label{gateandexpert}
\end{eqnarray}

In MCGSM, Markov assumption was made and the past context $\mathbf{x}_{<ij}$  was actually limited to a small causal neighborhood. However natural images exhibit long range correlations and any smaller neighbourhood fails to capture them. On the other hand increasing the neighbourhood leads to dramatic increase in number of parameters. In order to take into account such dependencies  \cite{theis2015generative} have proposed to use two dimensional Spatial Long Short Term Memory (LSTMs) \cite{graves2012neural} units for summarizing the causal context through their hidden representation $\mathbf{h}_{ij}$ at location $ij$ as, 
\begin{equation}
    \mathbf{h}_{ij}  = f(\mathbf{x}_{<ij}, \mathbf{h}_{i-1,j},  \mathbf{h}_{i,j-1})
\end{equation}
where $f$ is a complex non linear function with memory elements analogous to physical read, write and erase elements thus giving it the ability to model the long term dependencies in sequences. This formulation results in replacement of the finite context $\mathbf{x}_{<ij}$ in conditional modeling equation \eqref{mcgsm} with $\mathbf{h}_{ij}$, thus bringing in the summary of entire causal context. Thus, the complete model is specified as follows:
\begin{eqnarray}
  p(\mathbf{x}) &=& \prod_{ij} p(x_{ij}|\mathbf{h}_{ij}, \boldsymbol{\theta}) \label{ride-Factorization} \\
  p(x_{ij}|\mathbf{h}_{ij},\boldsymbol{\theta}) &=& \sum_{c,s} p(c,s|\mathbf{h}_{ij},\boldsymbol{\theta})p(x_{ij}|\mathbf{h}_{ij},c,s,\boldsymbol{\theta}), \label{ride}
\end{eqnarray}
Using Recurrent Image Density Estimator (RIDE)  \cite{theis2015generative} have achieved one of the state-of-the-art results in terms of log-likelihood scores. For more details we recommend the reader to go through \cite{theis2015generative}. 

\section{Compressive Image Recovery Using RIDE}
Here we consider the problem of image restoration from linearly compressed measurements $\mathbf{y} = A \mathbf{x} + \mathbf{n}$, where the linear transformation $A$ is a $M \times N$  with $M<N$, $\mathbf{n}$ is noise in the observation with known statistics. 
\subsection{MAP Inference via Backpropagation}
Sequential sampling of the conditional factors has been used by RIDE to generate image samples from the joint distribution  \cite{theis2015generative}. On similar lines, one method to do inference is to sample from the posterior distribution. But here sequential sampling is not possible and we have to resort to Markov Chain Monte Carlo methods such as Gibbs sampling which are computationally expensive even for smaller image sizes. Hence, we use Maximum-A-Posteriori principle to find the desired image $\mathbf{\hat{x}}$,
\begin{equation}
  \mathbf{\hat{x}}=arg \max_{\mathbf{x}}{p\left(\mathbf{x}|\mathbf{y}\right)}
  = arg \max_{\mathbf{x}}{p\left(\mathbf{x}\right)p\left(\mathbf{y}|\mathbf{x}\right)}
  \label{map}
\end{equation}

The prior term $p\left(\mathbf{x}\right)$ is specified by the generative model \eqref{ride-Factorization},\eqref{ride} and the likelihood is given by $p(\mathbf{y}|\mathbf{x})\propto exp(-||\mathbf{y}-A\mathbf{x}||^2/\sigma^2)$ for the isotropic Gaussian noise case. 

We apply gradient ascent to the net posterior distribution in order to obtain the reconstructed image. After log transforming the product in \eqref{map}, the gradient with respect to the prior is given by:

\begin{equation}
  \frac{\partial \log p(\mathbf{x})}{\partial x_{ij}} = \sum_{k \geq i, l \geq j} \frac{\partial \log p(x_{kl}|\mathbf{h}_{kl}, \boldsymbol{\theta})}{ \partial x_{ij}}
  \label{grad_xij}
\end{equation}

Due to the recurrent nature of the model, each pixel through its hidden representation can contribute to the likelihood of all the pixels that come after it in forward pass. In a similar fashion during backward pass the gradient from each pixel propagates to all the pixels prior to it in the sequence. Gradients with respect to log-likelihood are much easier to evaluate is given by:
\begin{equation}
    \nabla_{\mathbf{x}}  \log p(\mathbf{y|x})
   \propto 2A^T(\mathbf{y}-A\mathbf{x})
  \label{grad_xij}
\end{equation}
Using these gradient formulations, we can do gradient ascent for maximizing the log posterior with a momentum parameter for quick convergence. 
\begin{equation}
    \hat{\mathbf{x}}_{t+1} = \hat{\mathbf{x}}_{t} + \eta  \nabla_{\mathbf{x}} \log (p(\mathbf{x})p(\mathbf{y|x}))
  \label{grad_ascent}
\end{equation}
Where $\eta$ is the learning rate parameter.
\begin{figure}[t]
  \centering
   \begin{minipage}{0.2\textwidth}
   \centering
   \includegraphics[width=\textwidth]{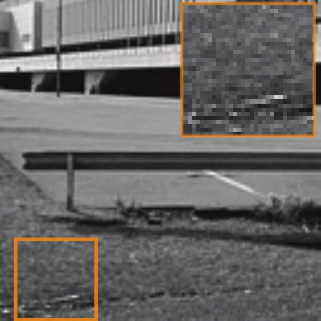}
   \centerline{Original Image}
   \end{minipage}
   \hspace{0.2cm}
   \begin{minipage}{0.2\textwidth}
   \centering
   \includegraphics[width=\textwidth]{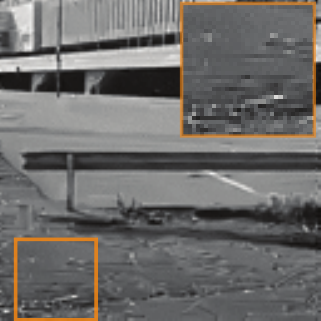}
   \centerline{w/o threshold, PSNR: 28.4 dB}
   \end{minipage}\\
   \vspace{0.3cm}
   \begin{minipage}{0.2\textwidth}
   \centering
   \includegraphics[width=\textwidth]{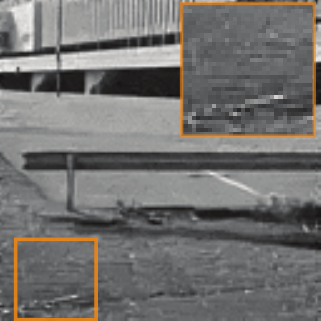}
   \centerline{w/ 3.5, PSNR: 29.90 dB}
   \end{minipage}
   \hspace{0.2cm}   
   \begin{minipage}{0.2\textwidth}
   \centering
   \includegraphics[width=\textwidth]{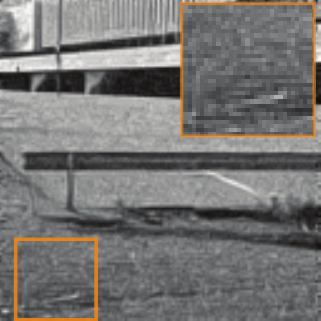}
   \centerline{w/ 3.0, PSNR: 26.97 dB}
   \end{minipage}   
    \caption{Compressive sensing image reconstructions from $30\%$ measurements obtained by varying entropy thresholds. The texture of the magnified patch is recovered better with the threshold.}
    \label{fig:entropy_threshold}
\end{figure}

\subsection{Tricks used for inference}
\subsubsection{Four directions}
Joint distribution \eqref{ride-Factorization} can be factorized in multiple ways, for example along each of the four diagonal directions of an image, i.e., top-right, top-left, bottom-right and bottom-left. Gradients from different factorizations are considered at each iteration of the inference, by flipping the image in the corresponding direction. This leads to faster convergence as compared to just considering one direction. While doing the inference on crops from randomly sampled BSDS test images, we observe that the convergence rate is roughly 2 times faster when considering four directions.

\begin{figure*}
\centering
\begin{minipage}{.20\textwidth}
\centerline{Original image}
\includegraphics[width=1\textwidth]{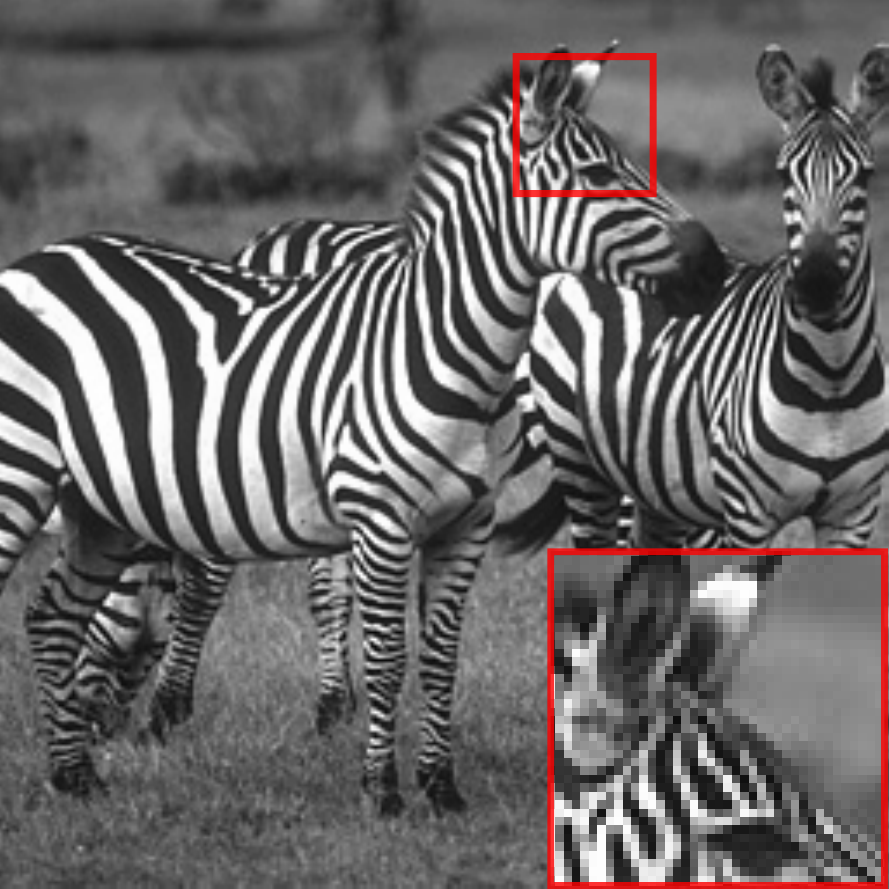}  
\centerline{}
\end{minipage}
\begin{minipage}{.20\textwidth}
\centerline{Masked image}
\includegraphics[width=1\textwidth]{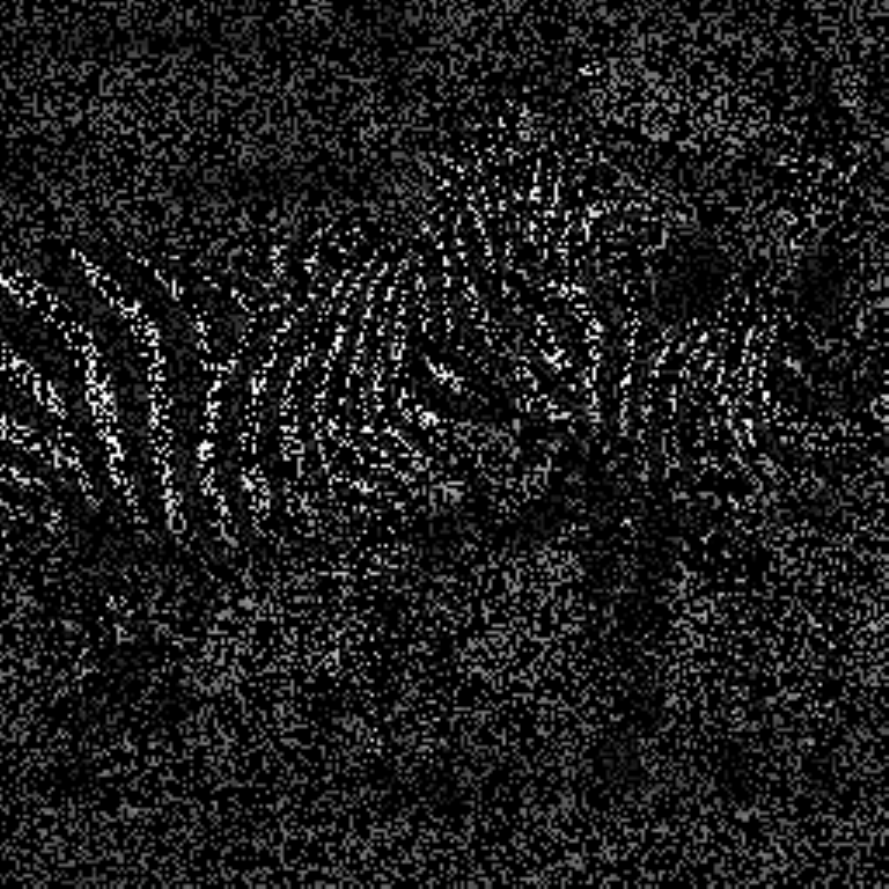}  
\centerline{}
\end{minipage}
\begin{minipage}{.20\textwidth}
\centerline{Multiscale KSVD}
\vspace{0.0845cm}
\includegraphics[width=1\textwidth]{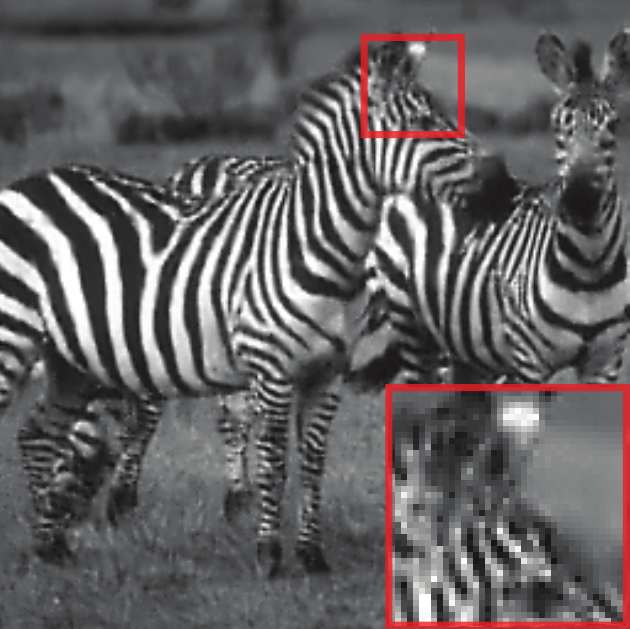}  
\centerline{21.21, 0.811}
\end{minipage}
\begin{minipage}{.20\textwidth}
\centerline{Ours}
\vspace{0.0845cm}
\includegraphics[width=1\textwidth]{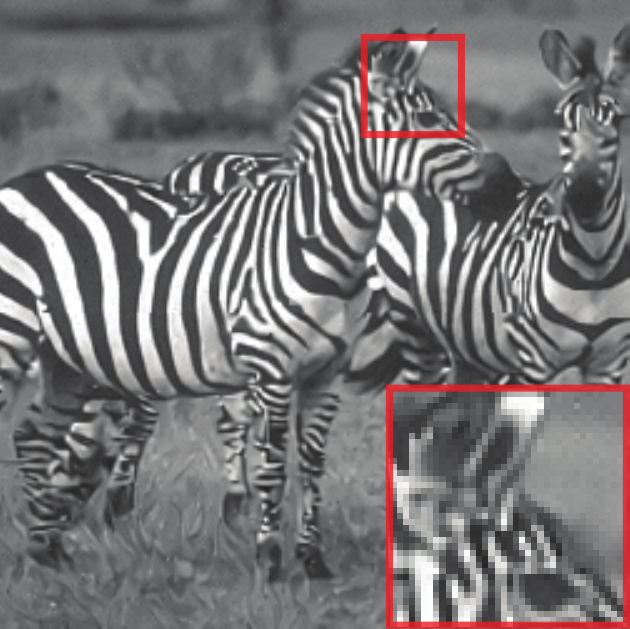}  
\centerline{ 22.07, 0.813 }
\end{minipage}

\caption{Inpainting comparisons: We compare our approach with the multiscale dictionary learning approach \cite{mairal2008learning}. Our method is able to recover the sharp edges better than the multiscale KSVD approach, as is evident in the zoomed region around zebra's eye. This is because our method is a global prior as compared to the patch-based multiscale KSVD  approach. The numbers mentioned below the figures are PSNR(left) and SSIM(right)}
\label{InpaintExpt}
\end{figure*}

\subsubsection{Entropy-based Thresholding} \label{ent_threshold}
While solving the MAP optimization, we observed that we can recover the edges quite well but texture regions are blurred. This happens because the RIDE model may not have the right mixture component (see \eqref{ride}) to explain the latent texture. In such cases, all the mixture components can be chosen with almost uniform probability, resulting in blurred texture. To detect such cases, in each iteration, we consider the posterior probability of scales and components in RIDE at each point as a metric to understand how confident the model is in modeling the distribution at that  point. This is evaluated through posterior entropy given as, 
\begin{equation}
    H(i,j) = -\sum_{c,s} p(c,s|\mathbf{x}_{<ij},x_{ij}) \log( p(c,s|\mathbf{x}_{<ij},x_{ij}) )
    \label{entropy}
\end{equation}
If the point lies on an edge, the posterior entropy is low as there are only certain selected components which can explain that edge. Whereas, if the point lies in a flat or textured patch, the posterior entropy is high and the point is equi-probable to come from different components and scales. Therefore, to reduce blurring we maintain a threshold on posterior entropy above which we clip the gradients to zero. \ref{fig:entropy_threshold} shows the effect of entropy constraint on the texture reconstruction.

\subsection{Compressive Image Recovery}
To demonstrate the effectiveness of our method, we consider the problems of image inpainting and compressive sensing imaging  \cite{baraniuk2007compressive}. 
In image inpainting our goal is to recover the missing pixels from a randomly masked image. We estimate the missing pixels by maximizing the prior over missing pixels, keeping the observed pixels constant. This is done by updating the gradients for only missing pixels. We have used the above mentioned entropy based gradient thresholding to avoid blurring the texture region. For SPC, we formulated the MAP inference as, 
\begin{equation}    
  \mathbf{\hat{x}}  = arg \max_{\mathbf{x}}{p\left(\mathbf{x}\right)} ~~ s.t. ~~ \mathbf{y} = \Phi\mathbf{x}
  \label{csmap}
\end{equation}
 To optimize the above we use projected gradients method, where after each gradient update solution is projected back on to the affine solution space for $\mathbf{y}=\Phi \mathbf{x}$. Every $k$-th iteration consists of the following two steps.
\begin{eqnarray}
    \mathbf{\hat{x}}_{k} &=& \mathbf{x}_{k-1} + \eta\nabla_{\mathbf{x}_{k-1}}p\left(\mathbf{x}\right) \\
    \mathbf{x}_k &=& \mathbf{\hat{x}}_{k}-\Phi^T\left(\Phi\Phi^T\right)^{-1}\left(\Phi\mathbf{\hat{x}}_{k}-\mathbf{y}\right)
\end{eqnarray}
In our experiments we consider row orthonormalized $\Phi$ and the term $\left(\Phi\Phi^T\right)^{-1}$ reduces to identity matrix. 

For the noisy measurements $\mathbf{y}$ will not exactly satisfy the constraint $\mathbf{y}=\Phi \mathbf{x}$. So, we cannot enforce hard constraints using the projected gradient method. Hence, we instead apply soft constraints by adding the term $\lambda\|\mathbf{y}-\Phi \mathbf{x}\|$ to the cost function for gradient ascent.
\section{Experiments}
For training the RIDE model we have used publicly available Berkeley Segmentation dataset (BSDS300). Following the instincts from \cite{theis2015generative}, we trained the model with increasing patch size in each epoch. Starting with 8x8 patch we go till 22x22 in steps of 2 for 8 epochs. We used the code provided by authors of RIDE in caffe, available here\footnote{https://github.com/lucastheis/ride/}. We start with a very low learning rate (0.0001) and decrease it to half the previous value after every epoch. We used Adam optimization \cite{kingma2014adam} for training the model. We observe that models with more than one spatial LSTM layer don’t result in much of improvement for our tasks of interest. Hence, we proceed with a single layer RIDE model for all the inference tasks in this paper. Also, we have used entropy based gradient thresholding \ref{ent_threshold} with threshold 3.5, to avoid blurring the texture regions in all the experiments. In order to accommodate for boundary issues we remove a two pixel neighbourhood around the image for PSNR and SSIM calculation in all the experiments. For a fair comparison, we also do the same for the reconstructions of TVAL3 \cite{li2013efficient} and D-AMP \cite{metzler2014denoising}. 
\subsection{Image Inpainting}
For image inpainting, we randomly removed 70$\%$ of pixels and estimated them using aforementioned inference method. We compared our approach with the multiscale adaptive dictionary learning approach  \cite{mairal2008learning}, which is an improvement over the KSVD algorithm, see Figure \ref{InpaintExpt}. It is clear from the figure that our approach is able to recover the sharp edges better than the multiscale KSVD approach. This is because our method is based on global image prior as compared to the patch-based multiscale KSVD approach.

\subsection{Single Pixel Camera}
\begin{figure*}
\centering
\begin{minipage}{.14\textwidth}
\includegraphics[width=1\textwidth]{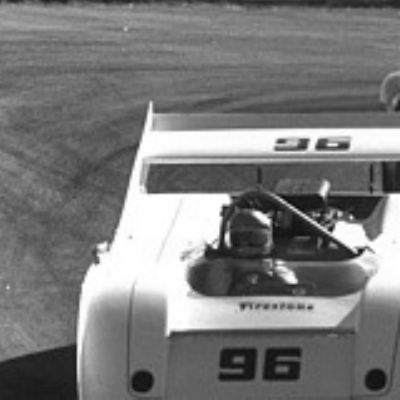}  
\centerline{car}
\end{minipage}\hspace{0.0cm}
\begin{minipage}{.14\textwidth}
\includegraphics[width=1\textwidth]{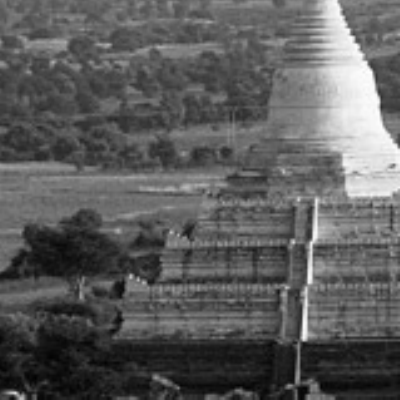}  
\centerline{monument}
\end{minipage}\hspace{0.0cm}
\begin{minipage}{.14\textwidth}
\vspace{0.08cm}
\includegraphics[width=1\textwidth]{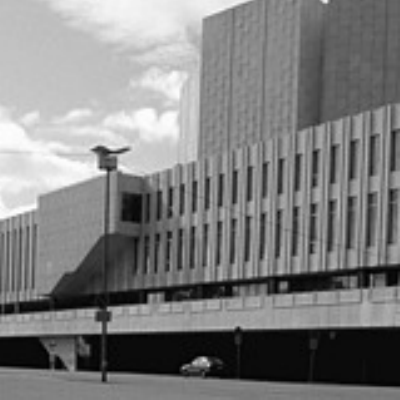}  
\centerline{building}
\end{minipage}\hspace{0.0cm}
\begin{minipage}{.14\textwidth}
\includegraphics[width=1\textwidth]{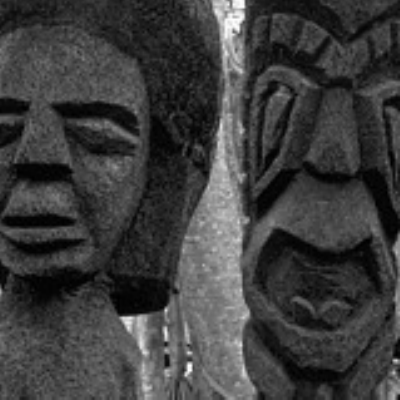}  
\centerline{statue}
\end{minipage}\hspace{0.0cm}
\begin{minipage}{.14\textwidth}
\includegraphics[width=1\textwidth]{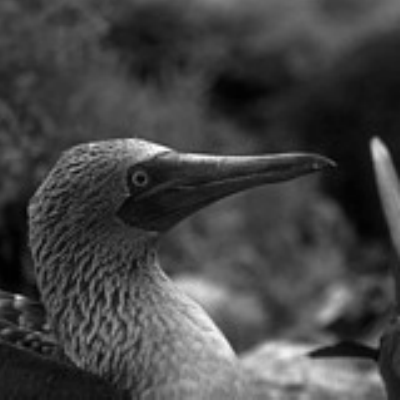}  
\centerline{bird}
\end{minipage}\hspace{0.0cm}
\caption{Randomly selected image crops of size 160x160 from BSDS300 test dataset used for CS reconstruction.}
\label{selImages}
\end{figure*}

In general, the SPC framework involves global multiplexing of the scene. But the recently proposed state-of-the-art methods for signal reconstruction, like ReconNet, are designed for local spatial multiplexing and can't handle the global multiplexing case directly. Our model, using Spatial LSTMs, can reason for long term dependencies in image sequences and is preferable for such kind of tasks. We show SPC reconstruction results on some randomly chosen images from the BSDS300 test set which were cropped to $160\times160$ size for computational feasibility, see Figure \ref{selImages}. We generate compressive measurements from them using random Gaussian measurement matrix with orthonormalized rows. We take measurements at four different rates $0.4$, $0.3$, $0.25$ and $0.15$. Using the projected gradient method, we perform gradient ascent for $300$ iterations for $0.4$, $0.3$ and $0.25$ measurement rates. For lower measurement rates, we run gradient ascent for $400$ iterations. Also, we follow the entropy thresholding procedure mentioned in section \ref{ent_threshold} with a threshold value of $3.5$ which we empirically found to be good for preserving textures.  In all the cases, we start with a random image uniformly sampled from $(0,1)$. Reconstruction results for five images are shown in Table \ref{CStable} and Figure \ref{CSexpt}. We were able to show improvements both in terms of PSNR and SSIM values for different measurement rates. Even at low measurement rates, our method preserves the sharp and prominent structure in the image. D-AMP has the tendency to over-smooth the image, whereas TVAL3 adds blotches to even the smooth parts of the image. 

\textbf{SPC with noise:}
To analyze the robustness of our framework with noise, we add different levels of Gaussian noise to the measurements obtained in the simulated case and obtain the reconstructions. The optimal value of $\lambda$ is empirically found out at different noise levels. Here we report our results in terms of average PSNR values over the same set of five images shown in Figure \ref{selImages} at different measurement rates. We can see that we are better than other methods at lower noise levels whereas at higher noise levels our performance drops slightly.

\textbf{Real Image Reconstruction:}
Here we consider the real measurements acquired from a single pixel camera using Fast Walsh Hadamard transform (FWHT) as $\phi$ matrix. Figure \ref{realim} depicts the reconstructions obtained in this case for the measurement rates of 15\%  and 30\%. It can be observed that our method provides superior reconstructions similar to the simulated case. Since we don't have original image here, we take reconstruction from D-AMP at 100\% measurements as the ground truth. Using this we evaluate the PSNR and SSIM metrics.

\section{Conclusions and Future Work}
We demonstrate that deep recurrent generative image models such as RIDE can be used effectively for solving compressive image recovery problems. The main advantages of using such models is that they are global priors and hence can model long term image dependencies. Also using the proposed MAP formulation we can solve many other image restoration tasks such as image deblurring, superresolution, demosaicing and computational photography problems such as coded aperture and exposure. 

\begin{table*}
\begin{center}  
\begin{tabular}{cccccccccc}
  \toprule
    \multirow{2}{*}{Figure Name}&\multirow{2}{*}{Method}&\multicolumn{2}{c}{M.R. = 40\%}&\multicolumn{2}{c}{M.R. = 30\%}&\multicolumn{2}{c}{M.R. = 25\%}&\multicolumn{2}{c}{M.R. = 15\%}\\
    \cline{3-10} & &  PSNR & SSIM & PSNR & SSIM & PSNR & SSIM & PSNR & SSIM \\
   \midrule
   
    \multirow{3}{*}{Car} & TVAL3  & 31.72 & 0.897 & 30.37 & 0.846 & 29.09	& 0.814 &  26.15 & 0.736	  \\
    & D-AMP & 34.00 & 0.908 & 32.31 & 0.877 & 30.05 & 0.839	 &  24.70 & 0.716	 \\
     & Ours & \textbf{36.05} & \textbf{0.932} & \textbf{34.24} & \textbf{0.901} &  \textbf{32.91} &  \textbf{0.868}	 &  \textbf{29.58} & \textbf{0.776}	  \\
      \midrule

    \multirow{3}{*}{Monument} & TVAL3  &28.10 & 0.796 & 28.43 & 0.750 & 27.69 & 0.710  & \textbf{26.13} & \textbf{0.611}	\\
     & D-AMP & 27.33 & 0.740 & 27.90 & 0.707 &27.19 & 0.665	 &	23.05 & 0.460	\\
     & Ours & \textbf{32.02} & \textbf{0.881}& \textbf{29.73} & \textbf{0.809} &  \textbf{28.78} & \textbf{0.766}	& 24.93 &  0.543	\\
      \midrule
       
    \multirow{3}{*}{Building} & TVAL3  & 28.40 & 0.842 & 26.16 & 0.784  & 25.13 &0.747 & 22.75 & 0.644	\\
     & D-AMP & \textbf{36.04} & \textbf{0.961} & 32.21 & 0.929 & 29.26 & 0.886 & 24.5 & 0.757	\\
     & Ours & 34.80 & 0.948 & \textbf{33.82} & \textbf{0.935} & \textbf{32.21} & \textbf{0.913}	 & \textbf{27.6} & \textbf{0.816}	\\
      \midrule
       
    \multirow{3}{*}{Statue} & TVAL3 & \textbf{28.01} & 0.777 & \textbf{26.67} & 0.712 & 26.08 & 0.675 & \textbf{24.59} & 0.583	\\
    & D-AMP & 26.90 & 0.661 & 25.80 & 0.613 & 25.20 & 0.586	& 22.86 & 0.455 \\
     & Ours & 27.97 & \textbf{0.805} & 26.59 & \textbf{0.742} & \textbf{26.12} & \textbf{0.711} &  24.14 & \textbf{0.599}	 \\
         \midrule
          
    \multirow{3}{*}{Bird} & TVAL3 & 32.57 & 0.901 & 31.75 & 0.874 & 30.68	&	0.847 & 28.30 & 0.771	\\
    & D-AMP & \textbf{38.45} & \textbf{0.970} & 31.54 & 0.874 &29.59 & 0.822	 & 24.98 & 0.688	\\
   & Ours & 37.70 & 0.948 & \textbf{35.19} & \textbf{0.922} & \textbf{33.52} & \textbf{0.892}	 & \textbf{29.3} & \textbf{0.786}	 \\
    \midrule
    \multirow{3}{*}{Mean} & TVAL3  &29.70 & 0.833 & 28.68 & 0.793 & 27.73	& 0.759 & 25.58 & 0.670\\
     & D-AMP & 32.54 & 0.848 & 29.95 & 0.800 & 28.26 & 0.760 & 24.02 & 0.615 \\
     & Ours & \textbf{33.71} & \textbf{0.903} & \textbf{31.91} & \textbf{0.862} & \textbf{30.71} & \textbf{0.830} & 	\textbf{27.11}	& \textbf{0.704}\\
   \bottomrule
   
\end{tabular}
\end{center}
\vspace{-0.2cm}
\caption{Comparisons of compressive imaging reconstructions at different measurement rates for the images shown in Figure \ref{selImages}. Our method outperforms the existing global prior based methods in most of the cases. }
\label{CStable}
\end{table*}

\begin{figure}[h]
    \centering
    \includegraphics[trim=50 225 2 10,clip,width=0.5\textwidth]{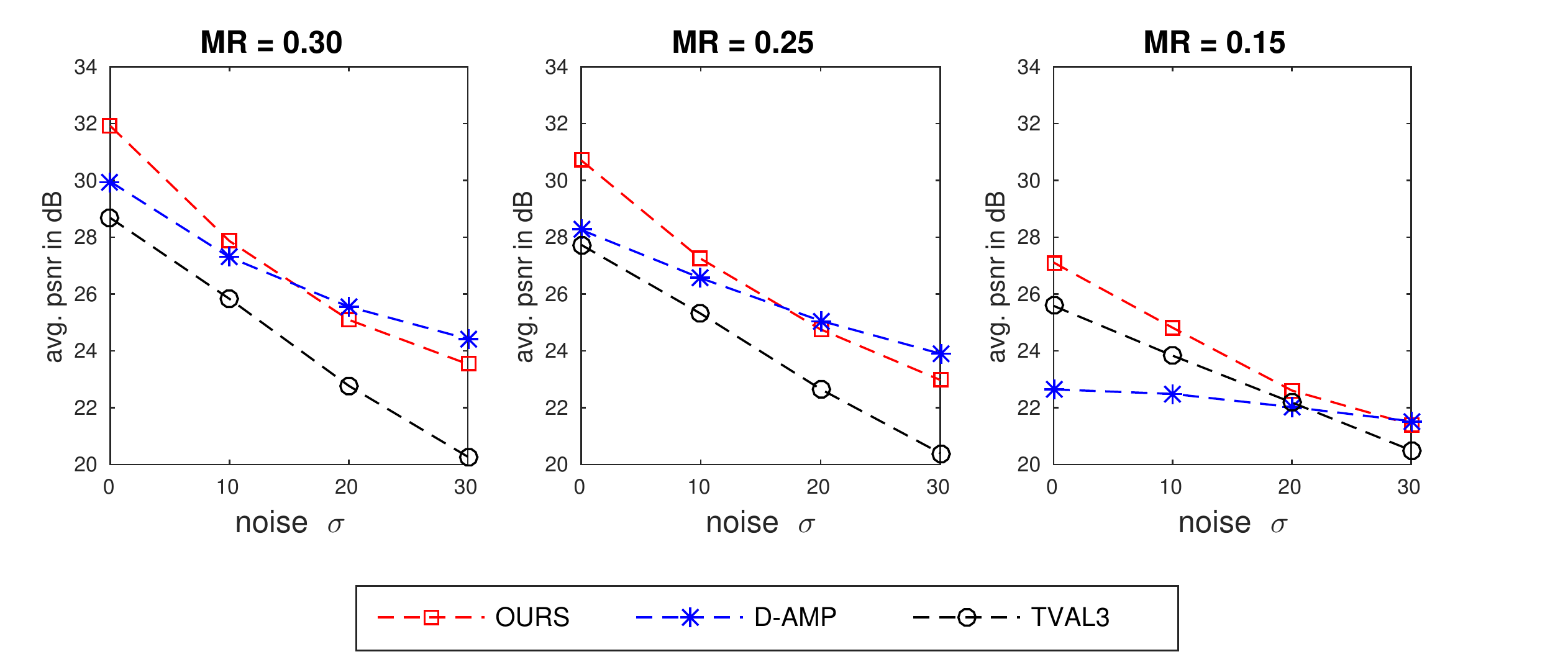}
    \caption{Performance of reconstructions from noisy measurements with different levels of Gaussian noise. (MR: Measurement Rate)}
    \label{noisy_cs}
\end{figure}

\begin{figure*}
\begin{minipage}{.01\textwidth}
\raggedleft
\begin{turn}{90}Reconstructions with 30\% M.R. \end{turn}
\end{minipage}
\begin{minipage}{.98\textwidth}
\centering
\begin{minipage}{.20\textwidth}
\centerline{Original image}
\vspace{0.01cm}
\includegraphics[trim=2 2 2 2,clip, width=1\textwidth]{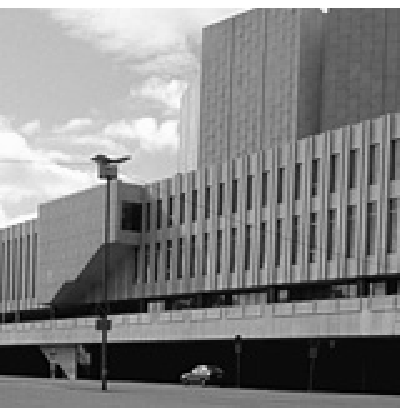}  
\centerline{}
\end{minipage}\hspace{0.1cm}
\begin{minipage}{.20\textwidth}
\centerline{TVAL3}
\vspace{0.1cm}
\includegraphics[trim=2 2 2 2,clip, width=1\textwidth]{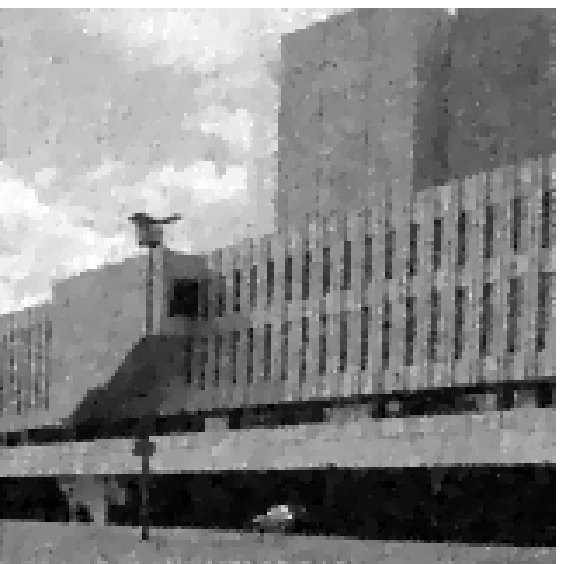}  
\centerline{26.16 dB, 0.784}
\end{minipage}\hspace{0.1cm}
\begin{minipage}{.20\textwidth}
\centerline{DAMP}
\vspace{0.1cm}
\includegraphics[trim=2 2 2 2,clip, width=1\textwidth]{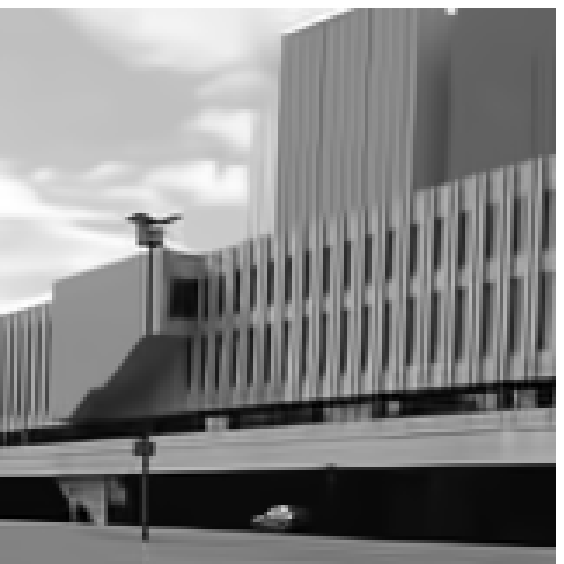}  
\centerline{32.21 dB, 0.929}
\end{minipage}\hspace{0.1cm}
\begin{minipage}{.20\textwidth}
\centerline{Ours}
\vspace{0.1cm}
\includegraphics[trim=2 2 2 2,clip, width=1\textwidth]{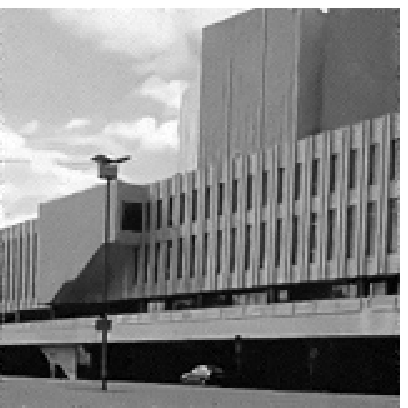}  
\centerline{33.82 dB, 0.935}
\end{minipage}\\

\begin{minipage}{.20\textwidth}
\vspace{0.04cm}
\includegraphics[trim=2 2 2 2,clip, width=1\textwidth]{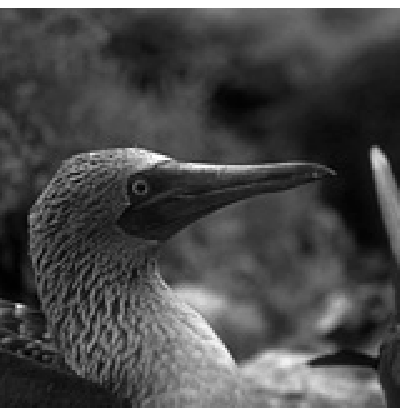}  
\centerline{}
\end{minipage}\hspace{0.1cm}
\begin{minipage}{.20\textwidth}
\vspace{0.1cm}
\includegraphics[trim=2 2 2 2,clip, width=1\textwidth]{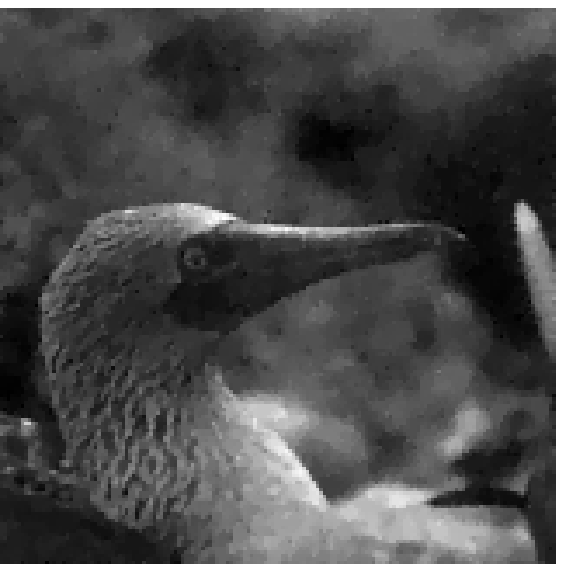}  
\centerline{31.75 dB, 0.874}
\end{minipage}\hspace{0.1cm}
\begin{minipage}{.20\textwidth}
\vspace{0.1cm}
\includegraphics[trim=2 2 2 2,clip, width=1\textwidth]{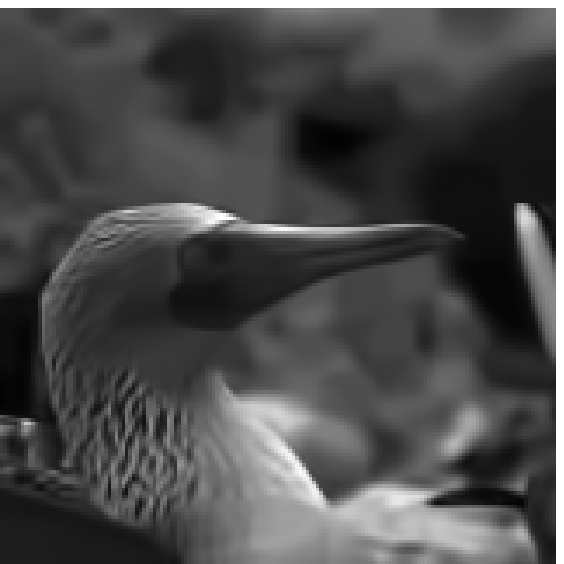}  
\centerline{31.54 dB, 0.874}
\end{minipage}\hspace{0.1cm}
\begin{minipage}{.20\textwidth}
\vspace{0.1cm}
\includegraphics[trim=2 2 2 2,clip, width=1\textwidth]{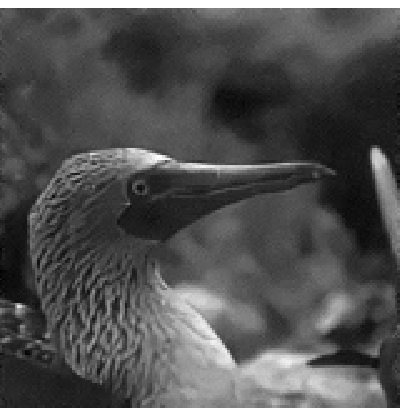}  
\centerline{35.19 dB, 0.922}
\end{minipage}
\end{minipage}

\centerline{}
\centerline{}

\begin{minipage}{.01\textwidth}
\raggedleft
\begin{turn}{90}Reconstructions with 15\% M.R. \end{turn}
\end{minipage}
\begin{minipage}{.98\textwidth}
\centering
\begin{minipage}{.20\textwidth}
\vspace{0.01cm}
\includegraphics[trim=2 2 2 2,clip, width=1\textwidth]{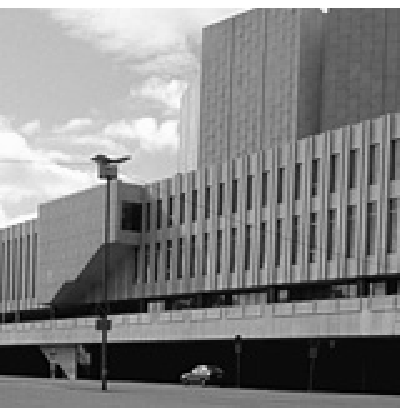}  
\centerline{}
\end{minipage}\hspace{0.1cm}
\begin{minipage}{.20\textwidth}
\vspace{0.1cm}
\includegraphics[trim=2 2 2 2,clip, width=1\textwidth]{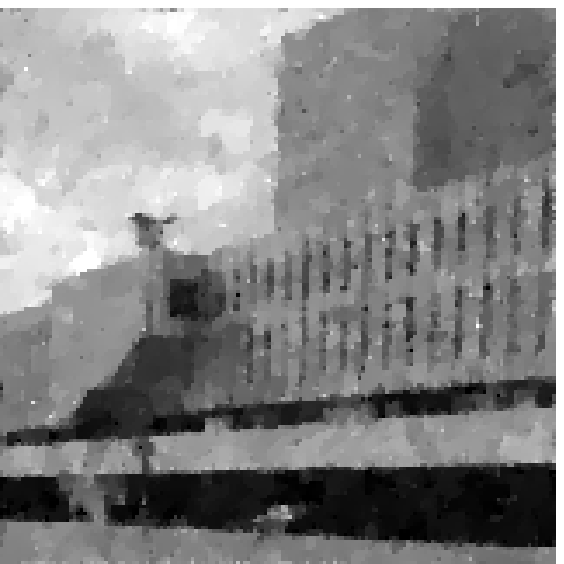}  
\centerline{22.75 dB, 0.644}
\end{minipage}\hspace{0.1cm}
\begin{minipage}{.20\textwidth}
\vspace{0.1cm}
\includegraphics[trim=2 2 2 2,clip, width=1\textwidth]{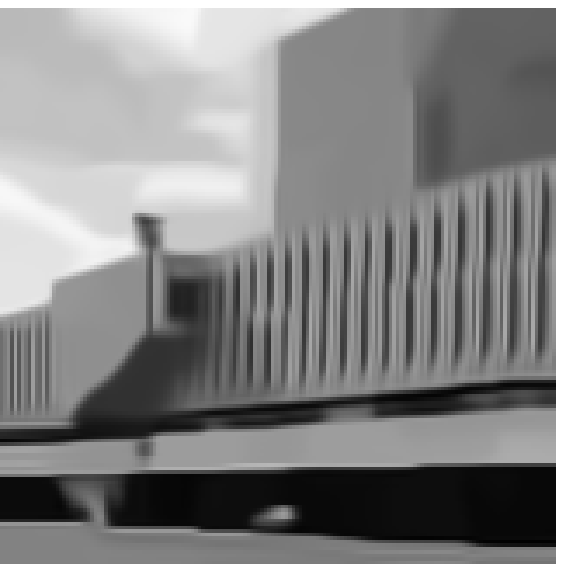}  
\centerline{24.5 dB, 0.757}
\end{minipage}\hspace{0.1cm}
\begin{minipage}{.20\textwidth}
\vspace{0.1cm}
\includegraphics[trim=2 2 2 2,clip, width=1\textwidth]{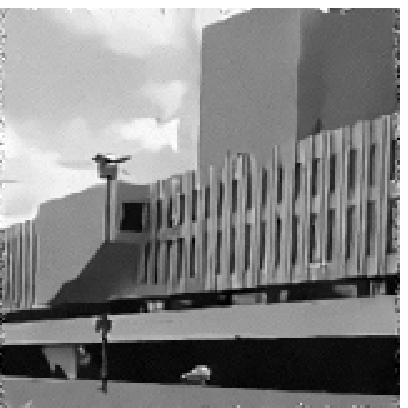}  
\centerline{27.6 dB, 0.816}
\end{minipage}

\begin{minipage}{.20\textwidth}
\vspace{-0.35cm}
\includegraphics[trim=2 2 2 2,clip, width=1\textwidth]{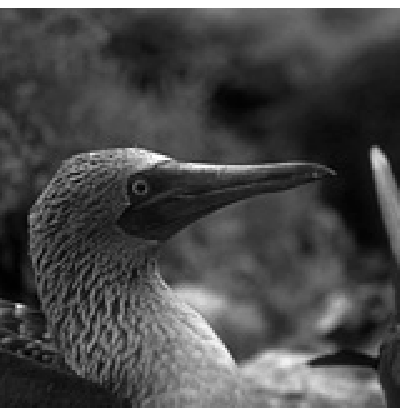}  
\end{minipage}\hspace{0.1cm}
\begin{minipage}{.20\textwidth}
\vspace{0.1cm}
\includegraphics[trim=2 2 2 2,clip, width=1\textwidth]{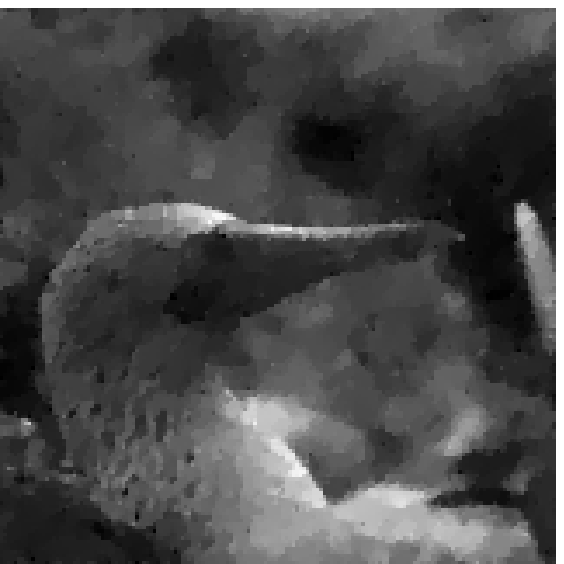}  
\centerline{28.30 dB, 0.771}
\end{minipage}\hspace{0.1cm}
\begin{minipage}{.20\textwidth}
\vspace{0.1cm}
\includegraphics[trim=2 2 2 2,clip, width=1\textwidth]{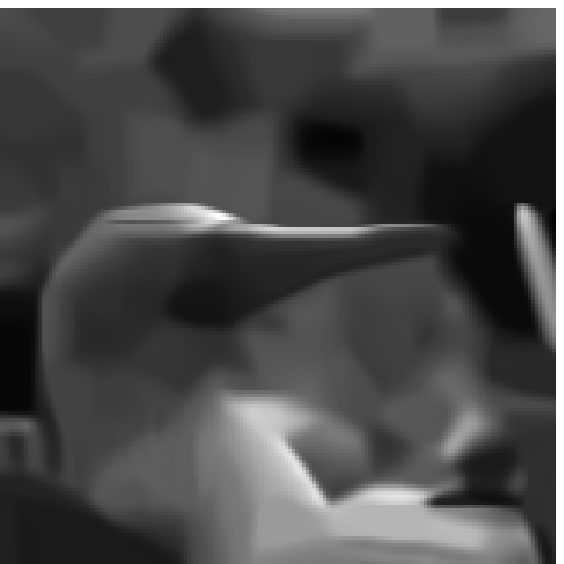}  
\centerline{24.98 dB, 0.688}
\end{minipage}\hspace{0.1cm}
\begin{minipage}{.20\textwidth}
\vspace{0.1cm}
\includegraphics[trim=2 2 2 2,clip, width=1\textwidth]{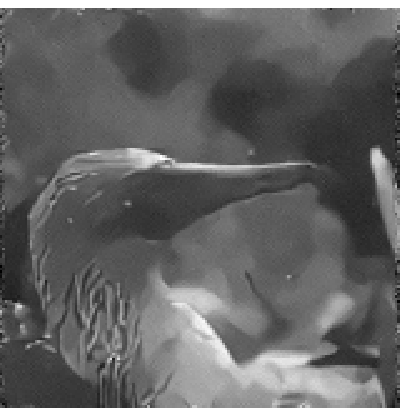}  
\centerline{29.30 dB, 0.786}
\end{minipage}
\end{minipage}
\caption{Images obtained by reconstruction from compressive measurements using D-AMP, TVAL3 and our method. Even at low measurement rates, our method preserves the sharp and prominent structures in the image. D-AMP has the tendency to over-smooth the image, whereas TVAL3 adds blotches to even the smooth parts of the image.}
\label{CSexpt}
\end{figure*}

\begin{figure*}
\begin{minipage}{.02\textwidth}
\begin{turn}{90}Reconstruction @ 0.3 M.R\end{turn}
\end{minipage}
\begin{minipage}{.98\textwidth}
\centering
\begin{minipage}{.21\textwidth}
\centerline{Full reconstruction}
\vspace{0.08cm}
\includegraphics[trim=2 2 2 2,clip, width=1\textwidth]{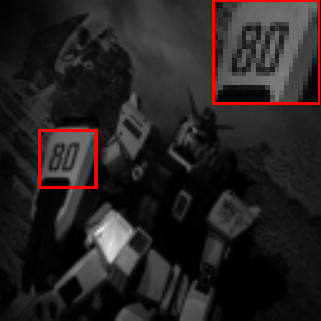}  
\centerline{}
\end{minipage}\hspace{0.1cm}
\begin{minipage}{.21\textwidth}
\centerline{TVAL3}
\vspace{0.08cm}
\includegraphics[trim=2 2 2 2,clip, width=1\textwidth]{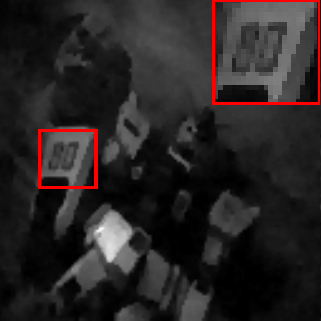}  
\centerline{32.31 dB, 0.876}
\end{minipage}\hspace{0.1cm}
\begin{minipage}{.21\textwidth}
\centerline{DAMP}
\vspace{0.1cm}
\includegraphics[trim=2 2 2 2,clip, width=1\textwidth]{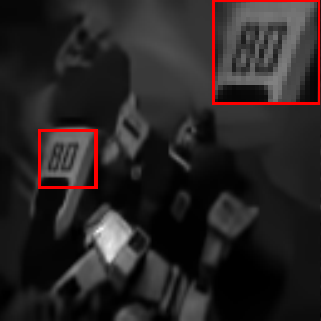}  
\centerline{32.95 dB, 0.883}
\end{minipage}\hspace{0.1cm}
\begin{minipage}{.21\textwidth}
\centerline{Ours}
\vspace{0.1cm}
\includegraphics[trim=2 2 2 2,clip, width=1\textwidth]{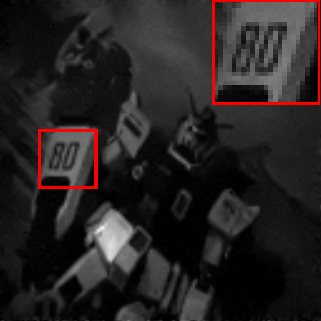}  
\centerline{35.87 dB, 0.908}
\end{minipage}
\end{minipage}

\centerline{}
\centerline{}

\begin{minipage}{.02\textwidth}
\begin{turn}{90}Reconstruction @ 0.15 M.R\end{turn}
\end{minipage}
\begin{minipage}{.98\textwidth}
\centering
\begin{minipage}{.21\textwidth}
\vspace{0.04cm}
\includegraphics[trim=2 2 2 2,clip, width=1\textwidth]{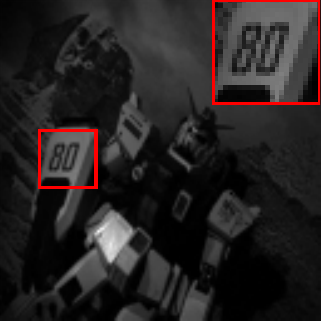}  
\centerline{}
\end{minipage}\hspace{0.1cm}
\begin{minipage}{.21\textwidth}
\vspace{0.1cm}
\includegraphics[trim=2 2 2 2,clip, width=1\textwidth]{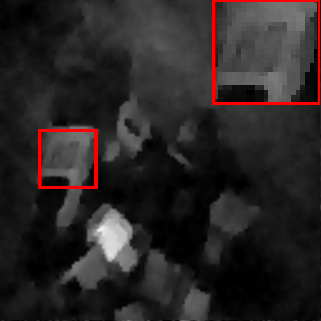}  
\centerline{28.34 dB, 0.760}
\end{minipage}\hspace{0.1cm}
\begin{minipage}{.21\textwidth}
\vspace{0.1cm}
\includegraphics[trim=2 2 2 2,clip, width=1\textwidth]{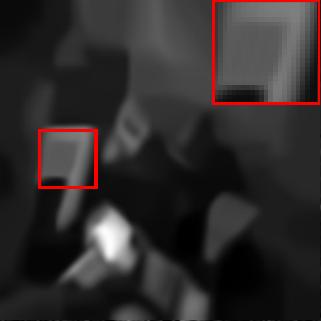}  
\centerline{24.68 dB, 0.687}
\end{minipage}\hspace{0.1cm}
\begin{minipage}{.21\textwidth}
\vspace{0.1cm}
\includegraphics[trim=2 2 2 2,clip, width=1\textwidth]{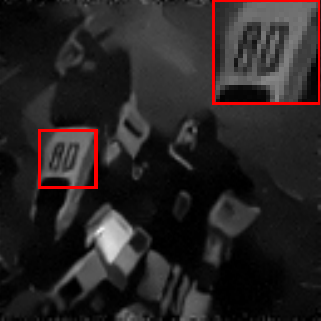}  
\centerline{31.12 dB, 0.813}
\end{minipage}
\end{minipage}
\caption{Reconstructions with 30\% measurement (top row) rate and 15\% measurement rate (bottom row) on real measurements acquired from single pixel camera.(Data courtesy: Dr. Aswin Sankaranarayanan) }
\label{realim}
\end{figure*}

{\small
\bibliographystyle{ieee}
\bibliography{egbib}
}

\end{document}